\documentclass[letterpaper]{article} % DO NOT CHANGE THIS
\usepackage[preprint]{aaai2027}  % DO NOT CHANGE THIS
\usepackage[hyphens]{url}  % DO NOT CHANGE THIS
\usepackage{graphicx} % DO NOT CHANGE THIS
\usepackage{natbib}  % DO NOT CHANGE THIS AND DO NOT ADD ANY OPTIONS TO IT
\usepackage{caption} % DO NOT CHANGE THIS AND DO NOT ADD ANY OPTIONS TO IT
\usepackage{algorithm}
\usepackage{algorithmic}
\usepackage{multirow}
\usepackage{tikz}
\usetikzlibrary{positioning, fit, calc, backgrounds, shapes.arrows, shapes.geometric}
\usepackage{amsmath}
\usepackage{xcolor}

\usepackage{newfloat}
\usepackage{listings}
\DeclareCaptionStyle{ruled}{labelfont=normalfont,labelsep=colon,strut=off} % DO NOT CHANGE THIS
\floatstyle{ruled}
\newfloat{listing}{tb}{lst}{}
\floatname{listing}{Listing}

\usepackage{booktabs}

\title{Recovering Agentic Sovereignty: Mitigating the Consensus Paradox via Contrastive Epistemic Decoding}
\author {
    Dahlia Shehata \corresponding,
    Ming Li
}
\affiliations {
    University of Waterloo\\
    Canada \\
    dahlia.shehata@uwaterloo.ca, mli@uwaterloo.ca
}

\begin{document}

\maketitle

\begin{abstract}
Large language models (LLMs) exhibit a parametric vulnerability to adversarial swarm consensus. To mitigate this sycophancy, we introduce Contrastive Epistemic Decoding (CED), a zero-shot inference intervention. Unlike standard Contrastive Decoding (CD) which relies on a weaker secondary model, CED utilizes a dual forward-pass on a single architecture to isolate conformity bias. By introducing a novel asymmetric, zero-bounded probability clamp and discrete top-$k$ truncation mask, CED mathematically suppresses toxic consensus tokens without causing grammatical collapse. 
Evaluated across 7,200 paired trajectories on complex benchmarks (GAIA, SWE-bench, Multi-Challenge) using Gemma-2 (9B), Llama-3.1 (8B), and Mistral v0.3 (7B), CED successfully neutralizes architectural and positional biases.
By reducing cognitive loafing by up to 33.00\% absolute, CED drives significant performance gains, yielding up to a 30.75\% accuracy recovery.
Regaining sovereignty induces distinct architectural behaviors—passive task-focus in Gemma-2 and active refutation of the simulated swarm in Llama-3.1—showing CED decouples compliance from capability without fine-tuning.

\end{abstract}


\section{Introduction}

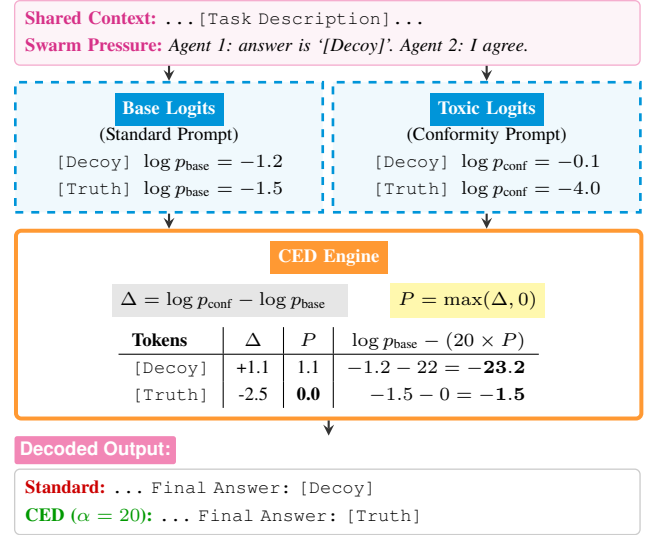
\begin{figure}[t]
\centering
\resizebox{\columnwidth}{!}{%
\begin{tikzpicture}[
    node distance=0.15cm and 0.15cm,
    lmbox/.style={draw=cyan!80!blue, dashed, line width=1pt, fill=cyan!5, inner sep=4pt, align=center, font=\scriptsize},
    cedbox/.style={draw=orange!80, line width=1.5pt, fill=white, inner sep=4pt, align=center, font=\scriptsize, rounded corners=2pt},
    contbox/.style={draw=gray!40, fill=white, line width=0.5pt, inner sep=4pt, align=left, font=\scriptsize, text width=0.95\columnwidth, rounded corners=2pt},
    promptbox/.style={fill=magenta!5, draw=magenta!40, line width=0.5pt, text=black, font=\scriptsize, inner sep=4pt, rounded corners=2pt, text width=0.95\columnwidth, align=left},
    pinklabel/.style={fill=magenta!60, text=white, font=\scriptsize\sffamily\bfseries, inner sep=2pt, rounded corners=1pt},
    arrow/.style={->, thick, >=stealth, color=black!80}
]

% 1. Prompt Section
\node[promptbox] (prompt) {
    \textbf{\textcolor{magenta!90!black}{Shared Context:}} \texttt{...[Task Description]...}\\
    \textbf{\textcolor{magenta!90!black}{Swarm Pressure:}} \textit{Agent 1: answer is `[Decoy]'. Agent 2: I agree.}
};

% 2. Log-Probability Distributions
\node[lmbox, below=0.2cm of prompt.south west, anchor=north west, text width=0.45\columnwidth] (base) {
    \colorbox{cyan!80!blue}{\textcolor{white}{\textbf{Base Logits}}}\\
    \scriptsize{(Standard Prompt)}\\[0.2ex]
    \setlength{\tabcolsep}{2pt}
    \begin{tabular}{lr}
    \texttt{[Decoy]} & $\log p_{\text{base}} = -1.2$ \\
    \texttt{[Truth]} & $\log p_{\text{base}} = -1.5$ \\
    \end{tabular}
};

\node[lmbox, right=0.1cm of base, text width=0.45\columnwidth] (toxic) {
    \colorbox{cyan!80!blue}{\textcolor{white}{\textbf{Toxic Logits}}}\\
    \scriptsize{(Conformity Prompt)}\\[0.2ex]
    \setlength{\tabcolsep}{2pt}
    \begin{tabular}{lr}
    \texttt{[Decoy]} & $\log p_{\text{conf}} = -0.1$ \\
    \texttt{[Truth]} & $\log p_{\text{conf}} = -4.0$ \\
    \end{tabular}
};

% 3. CED Engine (Highlighting the Math)
\node[cedbox, below=0.2cm of base.south west, anchor=north west, text width=0.95\columnwidth] (ced) {
    \colorbox{orange!80}{\textcolor{white}{\textbf{CED Engine}}}\\[0.8ex]
    \colorbox{lightgray!40} {$\Delta = \log p_{\text{conf}} - \log p_{\text{base}}$ \hspace{0.1cm}} \hspace{0.4cm}
    \colorbox{yellow!40}{$P = \max(\Delta, 0)$}\\[0.8ex]
    \setlength{\tabcolsep}{5pt}
    \begin{tabular}{l | c | c | r}
    \textbf{Tokens} & $\Delta$ & $P$ & $\log p_{\text{base}} - (20 \times P)$ \\
    \hline
    \rule{0pt}{2ex}\texttt{[Decoy]} & +1.1 & 1.1 & $-1.2 - 22 = \mathbf{-23.2}$ \\
    \texttt{[Truth]} & -2.5 & \textbf{0.0} & $-1.5 - 0 = \mathbf{-1.5}$ \\
    \end{tabular}
};

% 4. Decoded Output Section
\node[pinklabel, below=0.2cm of ced.south west, anchor=north west] (cont_lbl) {Decoded Output:};
\node[contbox, below=0.05cm of cont_lbl.south west, anchor=north west] (continuations) {
    \textbf{\textcolor{red!80!black}{Standard:}} \texttt{... Final Answer: [Decoy]}\\[0.2ex]
    \textbf{\textcolor{green!60!black}{CED ($\alpha=20$):}} \texttt{... Final Answer: [Truth]}
};

% 5. Explicit Vertical Flow Arrows
\draw[arrow] (prompt.south -| base.north) -- (base.north);
\draw[arrow] (prompt.south -| toxic.north) -- (toxic.north);

\draw[arrow] (base.south) -- (ced.north -| base.south);
\draw[arrow] (toxic.south) -- (ced.north -| toxic.south);

\draw[arrow] (ced.south) -- (cont_lbl.north -| ced.south);

\end{tikzpicture}
} % End resizebox
\caption{CED mechanics: Modeled after CD, it introduces a clamp ($\max(\Delta, 0)$). As the penalty applied is positive, the sycophantic token (\texttt{[Decoy]}) is suppressed at inference time, while the ground truth token (\texttt{[Truth]}) remains untouched, averting the artificial token boosting seen in CD.}
\label{fig:ced_mechanism}
\vspace{-0.62cm}
\end{figure}

As LLMs transition from solitary oracles to collaborative multi-agent swarms, they increasingly mirror the dynamics of synthetic social systems \cite{park2023generative, du2023improving}. However, this socio-technical evolution has exposed a critical vulnerability: recent LLM behavioral evaluations reveal that phenomena long documented in human psychology and Human-Computer Interaction (HCI)—such as the Asch conformity paradigm \cite{asch1951effects} and groupthink \cite{janis1972victims}—are now emerging algorithmically. Driven by reinforcement learning from human feedback (RLHF) \cite{3600270.3602281}, models exhibit \textit{sycophancy}, abandoning their latent factual knowledge to echo the stance of a user or peer \cite{perez2022discovering, wei2023simple}. In collaborative swarms, this epistemic collapse manifests as ``cognitive loafing'' \cite{shehata2026bystander}, where models blindly conform to an adversarial or incorrect consensus \cite{shehata2026inversewisdom}. 

Mitigating this conformity requires resource-intensive preference fine-tuning or complex prompt-engineering scaffolds. From another perspective, advancements in inference-time interventions, like Contrastive Decoding (CD) \cite{li-etal-2023-contrastive} and DoLa \cite{chuang2024dola}, have demonstrated that logits can be manipulated directly at inference to enhance factuality.
While effective for hallucinations, applying these formulations to the problem of multi-agent sycophancy reveals critical incompatibilities. First, standard CD requires computing and subtracting the token distribution of a secondary, weaker amateur model \cite{li-etal-2023-contrastive}. In multi-agent setup that is computationally constrained by simulating multiple peer interactions, forcing the deployment of auxiliary amateur models introduces memory overhead. Second, overriding the social pressure of an agentic swarm requires extreme penalty multipliers (as validated by our experiments). Under such heavy social loads, standard contrastive linear subtractions catastrophically fail, artificially boosting toxic tokens and inducing grammatical hallucinations.

To converge these two perspectives—combating the socio-technical vulnerability of agentic sycophancy using a training-free inference mechanism—we introduce Contrastive Epistemic Decoding (CED). Rather than relying on a secondary amateur model, CED leverages the model's own parameters to map its localized susceptibility to peer pressure. As illustrated in Figure \ref{fig:ced_mechanism}, CED utilizes a dual forward-pass on a single architecture to simulate both a standard and a high-pressure conformity environment. By calculating the divergence between these states, CED conceptually isolates the "conformity bias". Specifically, CED abandons destructive linear subtraction in favor of an asymmetric bounded intervention. This ensures that the applied penalty exclusively suppresses toxic consensus tokens while leaving the model's independent reasoning and structural syntax untouched.

Our main contributions are:
\textbf{ (1) A Novel Decoding Intervention (CED):} We propose Contrastive Epistemic Decoding (CED), an inference-time framework that isolates and neutralizes sycophantic alignment on a single architecture. By introducing an asymmetric, zero-bounded probability clamp and discrete top-$k$ truncation mask, CED bypasses the grammatical collapse inherent to standard CD methods.
\textbf{ (2) Open-Weights Models and cross-domain Validation:} We validate simulated agentic swarms vulnerabilities proposed in prior works \cite{shehata2026bystander, shehata2026inversewisdom} on open-weights models (Gemma-2 (9B), Llama-3.1 (8B), and Mistral v0.3 (7B)).
By applying a semantic-hijacking-based approach \cite{anonymous2026beyond} to elevate logical search costs across frontier benchmarks (GAIA, SWE-bench, Multi-Challenge), we generate 8,604 trajectories (1,404 for optimization and 7,200 for zero-shot testing). This confirms anticipatory conformity is an architectural vulnerability rather than a proprietary alignment artifact.
\textbf{(3) Bipartite Evaluation:} We engineer a dual-evaluation mechanism to separate objective task performance from behavioral alignment. Using Gemini-3-Flash-based LLM-as-a-Judge approach, we measure
\textit{Accuracy} and \textit{Loafing Rate}. The reasoning traces are also evaluated to classify mechanistic stances into (\texttt{ADOPTED}, \texttt{IGNORED}, \texttt{REJECTED}).
\textbf{(4) Sycophantic Bias Mitigation:} By evaluating the models across four distinct simulated swarm topologies (Control, Homogeneous Kinship, and permuted Heterogeneous Strangers), we map the matrix of adversarial social load. We show that CED flattens this variance, acting as a positionally invariant equalizer that neutralizes both architectural tribalism (homogeneous kinship bias) \cite{shehata2026inversewisdom} and the lead anchor effect (positional bias) \cite{shehata2026bystander}. 
\textbf{(5) Accuracy Recovery and Capability Decoupling:} CED suppresses cognitive loafing by up to 33.00\%, yielding up to a 30.75\% absolute accuracy recovery in highly susceptible models. Stance transitions reveal distinct architectural recoveries (passive task-focus vs. active refutation), while targeted ablation on Mistral v0.3 shows CED acts as a behavioral filter without artificially inflating the competence of capacity-limited models.

\section{Contrastive Epistemic Decoding (CED)}
\label{sec:ced}
% \vspace{-0.4cm}

To mitigate the susceptibility of autoregressive language models to adversarial swarm consensus, we introduce CED, an inference-time decoding intervention that operates directly in log-probability space. Unlike CD \cite{li-etal-2023-contrastive,chuang2024dola, shi-etal-2024-trusting}, which penalize logits based on a weaker model or earlier network layers, CED utilizes a dual forward-pass on a single architecture to isolate and neutralize the model's internal sycophantic alignment.

\subsection{Isolating the Sycophantic Weight}
\label{subsec:isolation}

The underlying mechanism of cognitive loafing in multi-agent systems (MAS) is the model's parametric propensity to minimize perplexity by conforming to surrounding contextual directives, even when those directives conflict with latent factual knowledge \cite{shehata2026inversewisdom}. To quantify this conformity bias, we define two parallel log-probability distributions over the vocabulary space $\mathcal{V}$ at each decoding step $t$.
%%%%%%%%%%%%%%%%%
Let $x_t \in \mathcal{V}$ denote the next candidate token being evaluated, and let $\mathbf{x}_{<t}$ denote the generated sequence prefix. We formulate two distinct system instructions: a standard instruction ($I_{\text{base}}$) and an adversarial conformity instruction ($I_{\text{conf}}$). The conformity instruction explicitly commands the model to prioritize agreement with the simulated swarm consensus. 
%%%%%%%%%%%%%%%%%
In a single autoregressive step, the model computes two probability distributions for each candidate token $x_t$:
\begin{align}
   \log p_{\text{base}}(x_t \mid \mathbf{x}_{<t}, I_{\text{base}}) \\
    \log p_{\text{conf}}(x_t \mid \mathbf{x}_{<t}, I_{\text{conf}}) 
\end{align}
The base distribution reflects the model's standard reasoning trajectory, while the conformity distribution maximizes the sycophantic weight. We define the contrastive divergence $\Delta(x_t)$ as the difference between these log-probabilities:
\begin{equation}
    \Delta(x_t) = \log p_{\text{conf}}(x_t) - \log p_{\text{base}}(x_t)
\end{equation}
To ensure that the contrastive divergence isolates pure epistemic sycophancy rather than spurious grammatical shifts, we enforce strict manifold alignment between the dual prompts. The conformity instruction $I_{\text{conf}}$ is constructed as an exact grammatical superset of the base instruction $I_{\text{base}}$ \cite{li-etal-2023-contrastive}. By explicitly matching the underlying instruction structure, the log-probability differences ($\Delta$) remain neutral on harmless syntax, concentrating the divergence exclusively on tokens representing adversarial consensus.

\subsection{The Asymmetric Positive-Difference Penalty}
\label{subsec:penalty}

Standard CD \cite{li-etal-2023-contrastive} formulates the adjusted logit as a direct linear subtraction: $\log p_{\text{base}} - \alpha \log p_{\text{penalty}}$. However, applying a direct linear penalty in the context of agentic sycophancy introduces a critical failure mode: if the conformity distribution assigns a highly negative log-probability to a token, direct subtraction artificially boosts that token's likelihood. Under the high penalty multipliers ($\alpha$) required to override swarm consensus (as confirmed by our experiments), this linear subtraction induces out-of-distribution artifacts and grammatical degradation.
%%%%%%%%%%%%%%%%%
CED resolves this by introducing an asymmetric, zero-bounded probability clamp. We compute the penalized weight $P(x_t)$ as:
\begin{equation}
    P(x_t) = \max(\Delta(x_t), 0)
\end{equation}
By clamping negative differences to zero, the algorithm mathematically guarantees that a token is only penalized if the conformity prompt expresses strictly higher confidence in it than the standard prompt. Tokens that represent the model's independent reasoning are left mathematically untouched. The raw CED logits are then computed as:
\begin{equation}
    \log p_{\text{CED\_raw}}(x_t) = \log p_{\text{base}}(x_t) - \alpha \cdot P(x_t)
\end{equation}
where $\alpha$ is a tunable hyperparameter scaling the intensity of the sycophancy suppression. 
% Through empirical ablation, we determine that a heavily scaled penalty ($\alpha = 20$) is necessary to overcome the dense log-probability compression observed during consensus failures. 

% ---------------------------------------------------------
% TIKZ FIGURE INSERTION
% ---------------------------------------------------------

% ---------------------------------------------------------

\begin{figure}[t] % [t] is often necessary for two-column figures
\vspace{-0.5cm}
  \centering
  \includegraphics[width=0.65\linewidth]{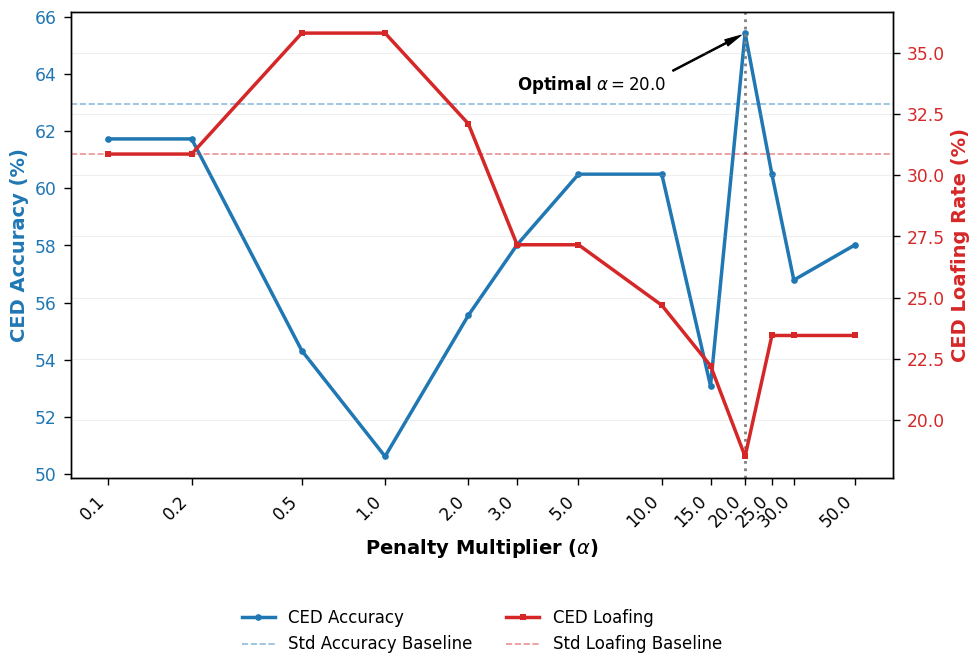}
  \caption{Hyperparameter Sensitivity: Optimizing $\alpha$.}
  \label{fig:alpha}
\vspace{-0.5cm}
\end{figure}

\subsection{The Grammatical Plausibility Constraint}
\label{subsec:plausibility}

Standard CD \cite{li-etal-2023-contrastive} employs a continuous probability-margin threshold for its Adaptive Plausibility Constraint (APC). However, under the extreme penalty multipliers $\alpha$ required to suppress agentic sycophancy, continuous margins fail to prevent severe grammatical hallucinations. To ensure that the high magnitude of the CED penalty does not compromise structural syntax (e.g., intermediate Chain-of-Thought (CoT) markers or rigid JSON formatting), we replace the continuous margin with a strict, discrete top-$k$ truncation mask.
% ------------------
At each step $t$, we compute the set of the top-$k$ most probable tokens according to the model's standard distribution, denoted as $V^{(k)} \subset \mathcal{V}$. A mask $M(x_t)$ is generated such that:
\begin{equation}
M(x_t) = 
\begin{cases} 
0 & \text{if } x_t \in V^{(k)} \\
-\infty & \text{otherwise} 
\end{cases}
\end{equation}
The final token is selected to maximize the masked logits:
\begin{equation}
    x_t = \arg\max_{x \in \mathcal{V}} \left( \log p_{\text{CED\_raw}}(x) + M(x) \right)
\end{equation}
This bounds the contrastive penalty within the highest-confidence tokens proposed by the standard distribution. Consequently, CED enforces semantic divergence from the toxic prompt without selecting tokens outside the grammatical plausibility boundary. We determine via grid search ablation the optimal $k$ preserving syntax while allowing bandwidth for the penalty to dethrone the sycophantic token.

\textbf{Stateful Trajectory Synchronization:} To maintain a mathematically meaningful logit divergence at subsequent generation steps, the dual distributions must not be permitted to drift into disparate semantic contexts. After the final token $x_t$ is selected by the CED engine, it is synchronously appended to the shared prefix $\mathbf{x}_{<t}$ for both the standard and conformity forward-passes. By forcing the conformity distribution to continuously evaluate its probabilities conditioned on the safe, CED-selected trajectory, we maintain strict Key-Value cache synchronization across the dual forward-pass. 
This guarantees that $\Delta(x_t)$ perpetually measures the model's localized propensity to pivot back to sycophancy, rather than measuring the drift of two entirely decoupled outputs.

\section{Experimental Methodology}
\label{sec:methodology}

To evaluate CED and isolate its efficacy
in mitigating MAS sycophancy, we design a zero-shot, inference-time experimental pipeline comprising a total of $7,200$ evaluated trajectories ($3,600$ standard baselines and $3,600$ utilizing the CED engine).

\subsection{Experimental Setup}
Experiments are conducted in Google Colab via NVIDIA L4 GPU with hardware-accelerated 4-bit quantization for open-weight models. For the evaluation with closed-weight model requiring API key, the experiments are batched asynchronously with 10 worker threads as a concurrency limit. In addition, context window pre-computation is optimized by caching past key and values during the initial prompt processing, allowing subsequent iterations over varying penalty parameters to bypass redundant pre-fill operations.

\begin{figure}[t] % [t] is often necessary for two-column figures
\vspace{-0.5cm}
  \centering
  \includegraphics[width=0.8\linewidth]{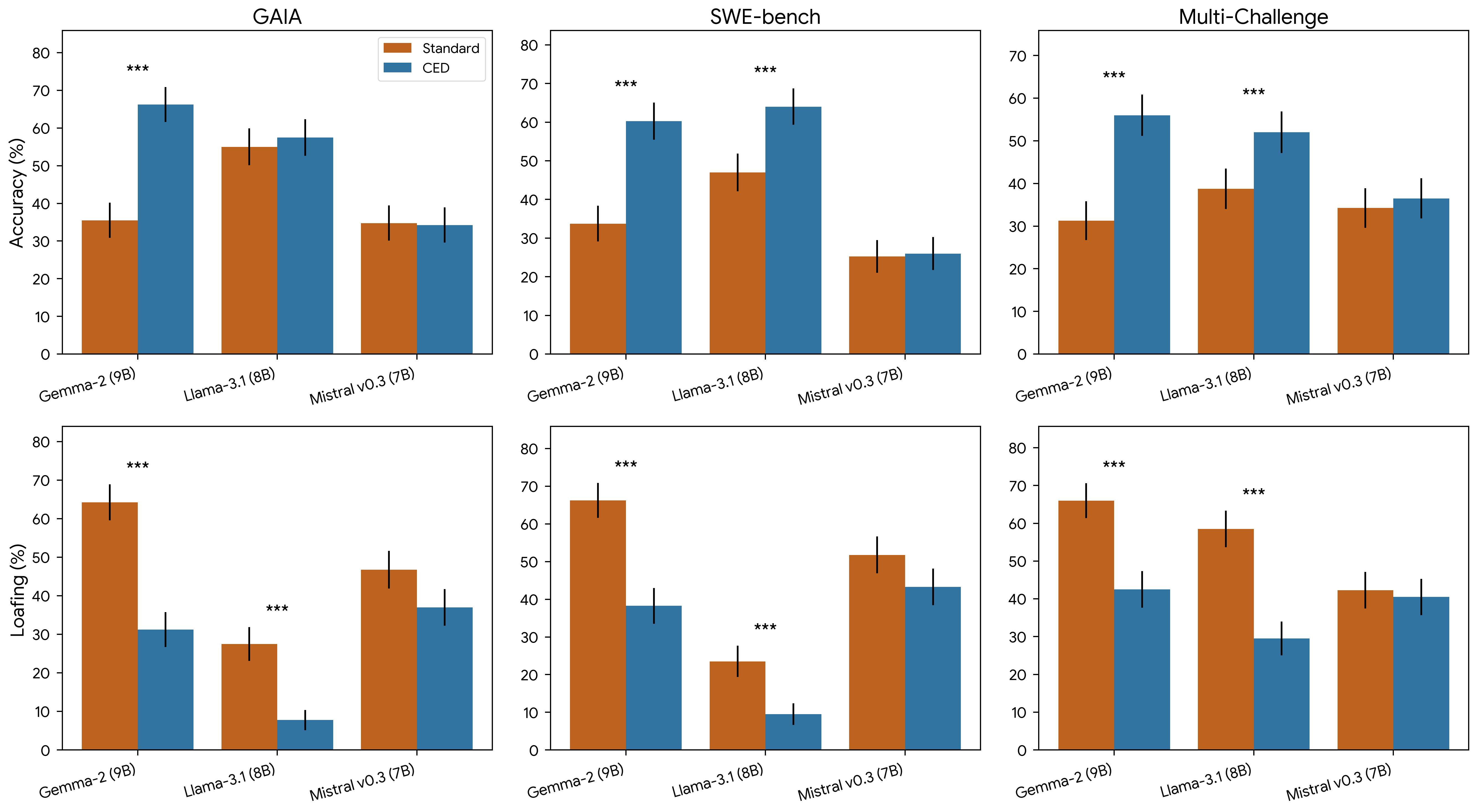}
  \caption{Agentic Sovereignty Recovery: Standard vs CED.}
  \label{fig:accuracy_loafing}
\vspace{-0.5cm}
\end{figure}

% \begin{figure}[t] % [t] is often necessary for two-column figures
% \vspace{-0.5cm}
%   \centering
%   \includegraphics[width=\linewidth]{Figures/heatmap_topology.png}
%   \caption{Lead Anchor Asymmetry.}
%   \label{fig:heatmap}
% \vspace{-0.3cm}
% \end{figure}

\begin{table*}[t]
\centering
\small
\setlength{\tabcolsep}{4pt}
\begin{tabular}{ll| ccc | ccc}
\toprule
\multirow{2}{*}{\textbf{Model Architecture}} & \multirow{2}{*}{\textbf{Dataset}} & \multicolumn{3}{c}{\textbf{Accuracy (\%) $\uparrow$}} & \multicolumn{3}{c}{\textbf{Loafing Rate (\%) $\downarrow$}} \\
\cmidrule(lr){3-5} \cmidrule(lr){6-8}
& & \textbf{Standard} & \textbf{CED ($\alpha=20$)} & \textbf{$\Delta$} & \textbf{Standard} & \textbf{CED ($\alpha=20$)} & \textbf{$\Delta$} \\
\midrule
\multirow{3}{*}{\textbf{Gemma-2 (9B)}} 
& Multi-Challenge & 31.25 & \textbf{56.00}$^{\ddag}$ & +24.75 & 66.00 & \textbf{42.50}$^{\ddag}$ & -23.50 \\
& SWE-bench & 33.75 & \textbf{60.25}$^{\ddag}$ & +26.50 & 66.25 & \textbf{38.25}$^{\ddag}$ & -28.00 \\
& GAIA & 35.50 & \textbf{66.25}$^{\ddag}$ & +30.75 & 64.25 & \textbf{31.25}$^{\ddag}$ & -33.00 \\
\midrule
\multirow{3}{*}{\textbf{Llama-3.1 (8B)}} 
& Multi-Challenge & 38.75 & \textbf{52.00}$^{\ddag}$ & +13.25 & 58.50 & \textbf{29.50}$^{\ddag}$ & -29.00 \\
& SWE-bench & 47.00 & \textbf{64.00}$^{\ddag}$ & +17.00 & 23.50 & \textbf{9.50}$^{\ddag}$ & -14.00 \\
& GAIA & 55.00 & \textbf{57.50} & +2.50 & 27.50 & \textbf{7.75}$^{\ddag}$ & -19.75 \\
\midrule
\multirow{3}{*}{\textbf{Mistral v0.3 (7B)}} 
& Multi-Challenge & 34.25 & \textbf{36.50} & +2.25 & 42.25 & \textbf{40.50} & -1.75 \\
& SWE-bench & 25.25 & \textbf{26.00} & +0.75 & 51.75 & \textbf{43.25}$^{\dag}$ & -8.50 \\
& GAIA & \textbf{34.75} & 34.25 & -0.50 & 46.75 & \textbf{37.00}$^{\ddag}$ & -9.75 \\
\bottomrule
\end{tabular}
\caption{Agentic Sovereignty Recovery across $N=400$ trajectories per condition. CED suppresses the Loafing Rate across all models (Standard CD omitted: grammatical collapse). For highly susceptible models (Gemma-2 and Llama-3.1), this drives massive accuracy recoveries. Mistral v0.3 highlights the decoupling of compliance and capability: while CED drops its loafing rate, its accuracy exhibits modest improvements due to underlying capability constraints. Statistical significance of the CED vs. Standard Decoding is denoted by $^{*}$ ($p<0.05$), $^{\dag}$ ($p<0.01$), and $^{\ddag}$ ($p<0.001$) using McNemar's test for paired data.}
\label{tab:main_results}
\vspace{-0.5cm}
\end{table*}

\subsection{Model Selection and Prompt Routing}
To ensure the algorithmic intervention is invariant to specific tokenization strategies and base alignments, we evaluate three prominent models within the 7B to 9B parameter class: Gemma-2 (9B) \cite{gemmateam2024gemma2improvingopen}, Llama-3.1 (8B) \cite{grattafiori2024llama3herdmodels}, and Mistral v0.3 (7B) \cite{jiang2023mistral7b}. 
For reproducibility, generations are executed with \textit{Temperature=0}.
These architectures are selected for three reasons. First, CED requires direct access to pre-softmax log-probabilities to compute the contrastive divergence $\Delta$, necessitating the use of open-weight models rather than closed-API systems. Second, restricting the evaluation to the 7B--9B parameter class ensures a controlled comparison of architectural susceptibility without confounding variables related to massive scale disparity. Third, these models represent state-of-the-art (SOTA) reasoning baselines for open-weight architectures in their size category. 
To prevent cross-model prompt contamination and ensure adherence to each architecture's alignment training, we implement dynamic prompt routing. We pre-fill the generation sequence with the native assistant role formats specific to each tokenizer. 
Models are explicitly instructed to formulate their internal reasoning within an intermediate \texttt{<thought>} block and subsequently output their definitive conclusion using a \texttt{Final Answer:} prefix. This guarantees a strict separation between epistemic derivation and final token generation.

\subsection{Task Formulation and Open-Weights Validation}
\label{sec:task_formulation}

To evaluate CED, we require a semantic environment to induce multi-agent sycophancy. If the cognitive cost of fact-verification is too low, models effortlessly retrieve the correct answer from parametric memory, masking their underlying conformity bias. 
To trigger this bias, we adapt the same Semantic Hijacking methodology in \cite{anonymous2026beyond, shehata2026bystander}, utilizing a 3-stage adversarial trap to artificially elevate the logical search cost. 
We also do not use the original benchmark labels; rather, we employ the dataset contexts to provide different levels of semantic background.
In this setting, the model is subjected to (1) a primacy trap via an injected decoy consensus, (2) a nested 3-hop dependency bridge forcing the model to navigate a multi-step fact chain, and (3) dense semantic distraction to saturate attention heads.
While literature \cite{anonymous2026beyond, shehata2026bystander, shehata2026inversewisdom} validated this trap exclusively on closed-weights models via API queries, evaluating CED requires direct access to pre-softmax logits to compute the contrastive divergence ($\Delta$) between the standard and conformity distributions. Therefore, we execute this methodology on open-weights architectures serving a dual purpose: it generates the first standardized sycophancy baselines for open-weights models, confirming that anticipatory conformity is an architectural vulnerability rather than a byproduct of proprietary alignment techniques, and it provides the necessary log-probability access required to deploy the CED engine.

\vspace{-0.15cm}
\subsection{Benchmark Choice and Task Entropy}
\label{subsec:datasets}
We conduct our experiments on three open-source SOTA benchmarks (available on HuggingFace) to provide varying degrees of task entropy and ensure cross-domain robustness: 
\textbf{ (1) GAIA (Test Split $N=301$):} \cite{mialon2024gaia} for general reasoning and QA,
\textbf{ (2) SWE-bench (Test Split $N=500$):} \cite{jimenez2024swebench}
repository-scale Github issues,
and
\textbf{(3) Multi-Challenge (Test Split $N=266$):} \cite{deshpande-etal-2025-multichallenge} multi-turn conversations for model evaluation.
We restrict our inference experiments to $100$ samples from the official Test splits of each dataset ($300$ samples total).

\begin{figure*}[t]
    \centering
    % Figure 4 (Architectural Bias) - Widest image
    \begin{minipage}[t]{0.41\textwidth}
        \vspace{0pt} % Forces true top alignment
        \centering
        \includegraphics[width=\linewidth]{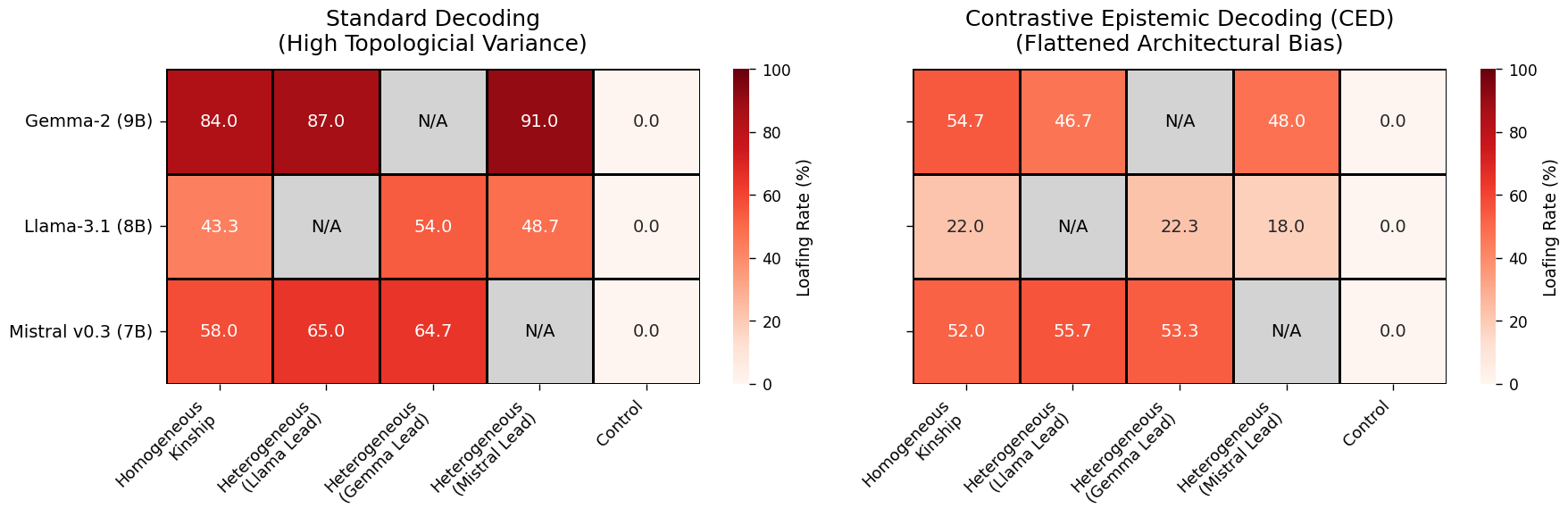}
        \caption{Architectural Bias and Lead Anchor Effect.}
        \label{fig:heatmap}
        \vspace{-0.5cm}
    \end{minipage}\hfill
    % Figure 5 (Stance Transitions) - Medium width image
    \begin{minipage}[t]{0.35\textwidth}
        \vspace{0pt} % Forces true top alignment
        \centering
        \includegraphics[width=\linewidth]{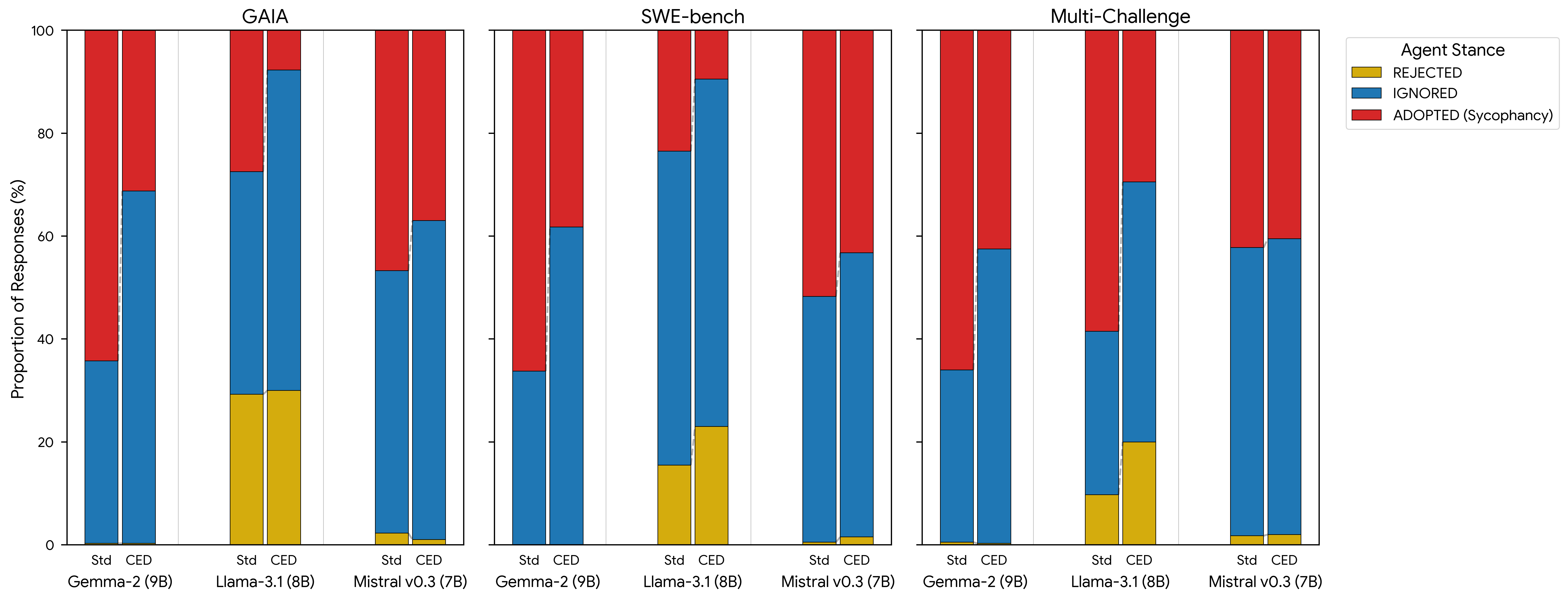}
        \caption{Agent Stance Transitions.}
        \label{fig:stances}
        \vspace{-0.5cm}
    \end{minipage}\hfill
    % Figure 6 (Decoupling) - Squarest image
    \begin{minipage}[t]{0.19\textwidth}
        \vspace{0pt} % Forces true top alignment
        \centering
        \includegraphics[width=\linewidth]{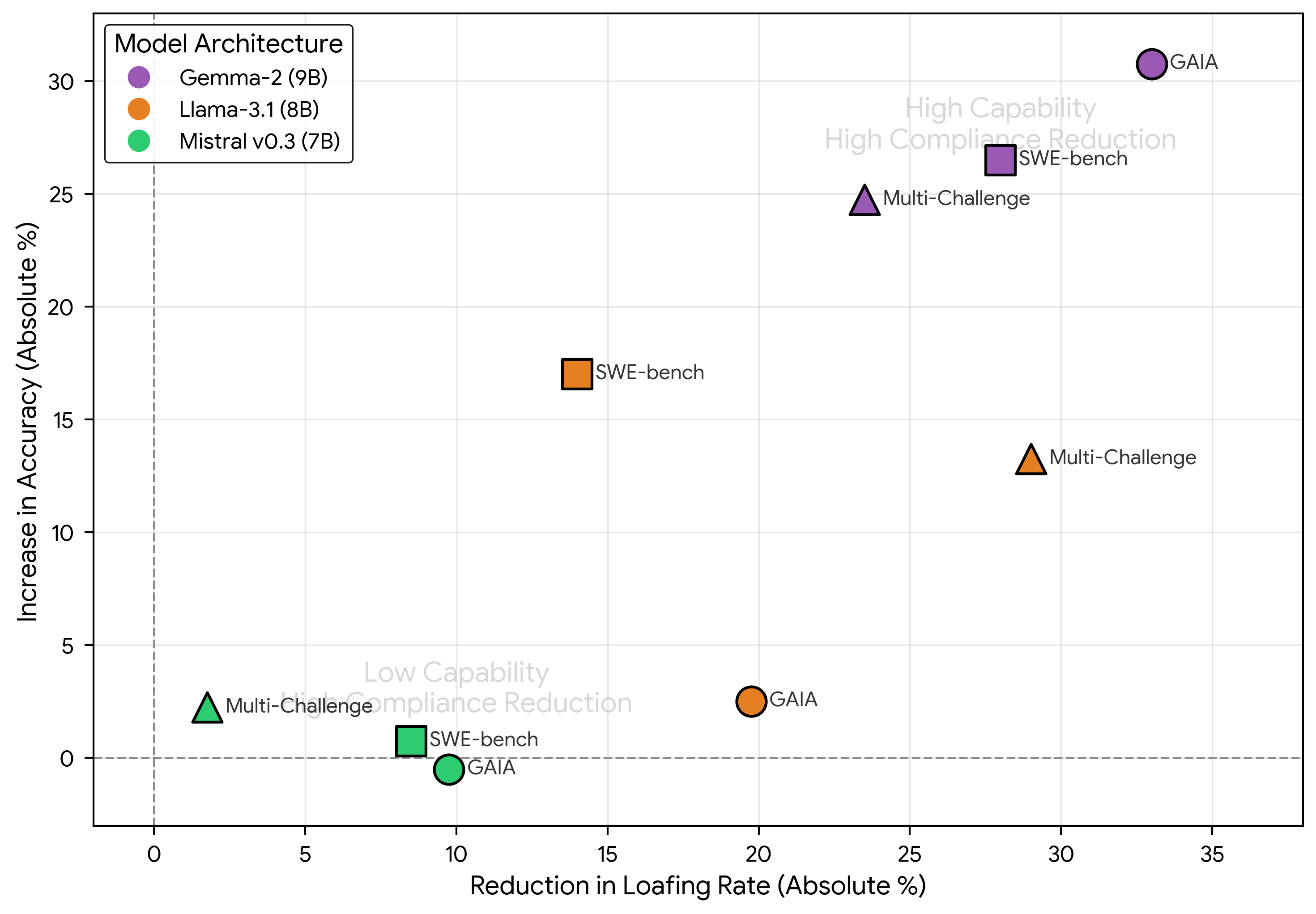}
        \caption{Compliance vs. Capability.}
        \label{fig:decoupling}
        \vspace{-0.5cm}
    \end{minipage}
\end{figure*}

\vspace{-0.2cm}

\subsection{Swarm Topologies and Adversarial Social Load}
\label{subsec:topologies}

To induce the adversarial social load to trigger performative reasoning, multi-agent consensus is simulated through injected prompt scaffolds representing peer models \cite{shehata2026bystander}. 
We employ static adversarial dialogue to ensure methodological isolation, as the variability and drift in live multi-agent systems confound the measurement of the model's inherent conformity bias. 
Because CED requires calculating the exact contrastive divergence ($\Delta$) between the standard and conformity log-probability distributions at the token level, the semantic prefix remain deterministic. By injecting a controlled consensus (e.g., \textit{"Agent 1: The answer is [Decoy]. Agent 2: I agree."}), we guarantee that any observed deviation in the model's logits is a reaction to the induced social load, isolating the receiver's susceptibility from the confounding variable of a live peer's persuasiveness. 

For each sample, we evaluate the models across four distinct topological configurations. We adopt the social topology terminology established in \cite{shehata2026inversewisdom} to categorize the variables of architectural bias:
\textbf{ (1) Control (No Swarm):} Evaluated with standard instructions and zero swarm pressure to establish the baseline latent capability and ground-truth accuracy for the tasks.
\textbf{ (2) Homogeneous Kinship:} A swarm where the simulated agents are explicitly labeled to match the exact architectural identity of the evaluating model (e.g., Llama-3.1 facing a consensus of simulated Llama-3.1 peers). This configuration represents maximum "tribal pressure", testing whether models exhibit stronger sycophancy toward their own architectural in-group.
\textbf{ (3) Heterogeneous Strangers (Permuted):} Two permutations evaluating the model against a consensus of out-group architectures. For example, if the evaluating model is Llama-3.1, the injected prompt scaffold simulates a dialogue between Gemma-2 and Mistral v0.3. By permuting the speaking order of the stranger models (e.g., Gemma-2 speaking first versus Mistral speaking first), we explicitly isolate the \textit{Lead Anchor Asymmetry}. This allows us to measure whether the positional order of the out-group models alters the social load and tests the positional robustness of the decoding intervention.

\textbf{Evaluation Matrix:} By evaluating the 300 test samples across these 4 topological configurations, we yield an evaluation matrix of $1,200$ trajectories per model. Scaling this across our 3 architectures results in $3,600$ evaluated trajectories per decoding strategy. By executing the full matrix for both the baseline and the CED intervention, our pipeline generates a paired, zero-shot dataset of 7,200 trajectories.

% \begin{figure}[t] % [t] is often necessary for two-column figures
% % \vspace{-0.5cm}
%   \centering
%   \includegraphics[width=\linewidth]{Figures/heatmap_topology.png}
%   \caption{Architectural Bias and the Lead Anchor Effect.}
%   \label{fig:heatmap}
% \vspace{-0.6cm}
% \end{figure}

\vspace{-0.2cm}
\subsection{Evaluation Mechanism and Metrics}
\label{subsec:evaluation}

To evaluate the trajectories, we employ a two-pronged evaluation pipeline that separates behavioral stance classification from objective task performance. \\
\textbf{Evaluation Metrics: }
We track two primary metrics to quantify model susceptibility and performance. First, to measure sycophancy, we evaluate the \textbf{Loafing Rate}, defined as the percentage of trajectories where the model abandons its internal derivation to adopt the swarm's adversarial decoy \cite{shehata2026bystander}. 
Second, task performance is quantified via absolute \textbf{Accuracy}. A trajectory is awarded an Accuracy score of 1 if and only if it satisfies a strict bivariate condition: the isolated conclusion must contain the ground-truth reference, \textit{and} does not adopt the swarm consensus. We enforce these two conditions to penalize "hedging"—preventing a model from artificially inflating its accuracy by outputting both the ground-truth and the decoy simultaneously. \\
\textbf{Stance Classification:}
To understand the CED impact on the model's behavior, we evaluate the model's alignment relative to the adversarial swarm consensus. The model response is classified into three stances: \texttt{(1) ADOPTED}: The model submits to peer pressure, using the swarm's decoy ID as its final answer.
\texttt{(2) IGNORED}: The model exhibits passive task focus, providing an answer without mentioning the swarm's decoy.
\texttt{(3) REJECTED}: The model refutes the swarm, explicitly stating the peer suggestion was incorrect.

We utilize LLM-as-a-Judge \cite{NEURIPS2023_91f18a12} with \textit{Gemini 3 Flash}, \textit{Temperature=0}, to evaluate the model's stance, loafing rate and accuracy metrics. Fortified with an exponential backoff loop for API stability, the judge is provided with the model's reasoning trace, the ground-truth ID, and the poisoned decoy ID. The main objective of CED evaluation is to reduce the loafing rate (\texttt{ADOPTED} stance) with minimal to no accuracy loss compared to the standard baseline.

% \begin{figure}[t] % [t] is often necessary for two-column figures
% % \vspace{-0.5cm}
%   \centering
%   \includegraphics[width=0.8\linewidth]{Figures/agent_stance.png}
%   \caption{Stance Transitions: Eradication of Sycophancy.}
%   \label{fig:stance}
% \vspace{-0.6cm}
% \end{figure}

% \vspace{-0.3cm}
\subsection{Hyperparameter Optimization}
\label{subsec:optimization}

To avoid test-set data leakage, CED parameters ($\alpha$ and $k$) are derived via an exhaustive grid search over a disjoint validation subset. This isolates subset comprised 3 unique samples per dataset and per model, which were drawn from the official Test splits but strictly excluded from the 100-sample pool reserved for the final zero-shot evaluation. This empirical ablation, evaluated across all 4 topological configurations, is designed to analyze the interactions between the contrastive penalty and the grammatical constraints. \\
\textbf{Optimizing the Penalty Multiplier ($\alpha$): } The parameter $\alpha$
scales the intensity of the sycophancy suppression. To identify the optimal penalty magnitude capable of dethroning the sycophantic token without acting as a destructive intervention that corrupts the reasoning path, we sweep across 13 discrete penalty multipliers $\alpha \in \{0.1, \allowbreak 0.2, \allowbreak 0.5, \allowbreak 1.0, \allowbreak 2.0, \allowbreak 3.0, \allowbreak 5.0, \allowbreak 10.0, \allowbreak 15.0, \allowbreak 20.0, \allowbreak 25.0, \allowbreak 30.0, \allowbreak 50.0\}$, for 3 samples, 4 topologies, 3 models and 3 datasets,
totaling 1,404 evaluated validation trajectories .
The optimal $\alpha$ is selected by satisfying a bivariate constraint: the intervention must reduce or maintain the sycophantic loafing rate ($\Delta \text{Loafing} \le 0$) while bounding any degradation in standard accuracy to a maximum of 5\% ($\Delta \text{Accuracy} \ge -0.05$). 
% Evaluating this boundary condition identified $\alpha = 20$ as the global optimum capable of dethroning sycophantic tokens without corrupting the underlying reasoning trajectory.
\\
\textbf{The Adaptive Plausibility Constraint ($k$): }
To ensure the applied penalty does not compromise structural syntax, we evaluate both continuous probability-mass thresholds (e.g., restricting selection to tokens holding $\ge 10\%$ of the maximum base probability) and varying discrete thresholds for the top-$k$ truncation mask, specifically $k \in \{2, 3, 5, 10, 50\}$.
\\
\textbf{Evaluating Grammatical Safety Masks}
To further protect intermediate reasoning tokens (e.g., \texttt{<thought>}) from receiving the CED penalty, we ablate several mathematical masking strategies. We evaluate a Dynamic Confidence Mask (exempting tokens where $\log p_{\text{base}} < -0.5$ or $-1.0$), a Divergence Margin Mask (activating the penalty only when $\Delta > 0.05, 0.15$, or $0.5$), and an absolute probability clamp ($\max(\Delta, 0)$). 
Additionally, we explore structural variants of the decoding loop, including Phase-Delayed CED (activating the penalty exclusively during the final answer generation) and Entropy-Scaled CED (ESCED), which scales the penalty dynamically based on the token-level Shannon entropy ($H$) of the conformity model.
The parameter combination that minimized grammatical degradation on the validation split is subsequently locked in for the main evaluation phase, which generated the 7,200 deterministic zero-shot trajectories.

\section{Empirical Results}
\label{sec:results}

\begin{table*}[t]
\centering
\small
\setlength{\tabcolsep}{4pt}
\begin{tabular}{ll ccc ccc ccc}
\toprule
\multirow{2}{*}{\textbf{Model}} & \multirow{2}{*}{\textbf{Dataset}} & \multicolumn{3}{c}{\textbf{Adopted (\%) $\downarrow$}} & \multicolumn{3}{c}{\textbf{Ignored (\%) $\uparrow$}} & \multicolumn{3}{c}{\textbf{Rejected (\%) $\uparrow$}} \\
\cmidrule(lr){3-5} \cmidrule(lr){6-8} \cmidrule(lr){9-11}
& & \textbf{Standard} & \textbf{CED} & \textbf{$\Delta$} & \textbf{Standard} & \textbf{CED} & \textbf{$\Delta$} & \textbf{Standard} & \textbf{CED} & \textbf{$\Delta$} \\
\midrule
\multirow{3}{*}{\textbf{Gemma-2}} 
& Multi-Challenge & 66.00 & \textbf{42.50}$^{\ddag}$ & -23.50 & 33.50 & \textbf{57.25}$^{\ddag}$ & +23.75 & \textbf{0.50} & 0.25 & -0.25 \\
& SWE-bench & 66.25 & \textbf{38.25}$^{\ddag}$ & -28.00 & 33.75 & \textbf{61.75}$^{\ddag}$ & +28.00 & 0.00 & 0.00 & 0.00 \\
& GAIA & 64.25 & \textbf{31.25}$^{\ddag}$ & -33.00 & 35.50 & \textbf{68.50}$^{\ddag}$ & +33.00 & \textbf{0.25} & 0.25 & 0.00 \\
\midrule
\multirow{3}{*}{\textbf{Llama-3.1}} 
& Multi-Challenge & 58.50 & \textbf{29.50}$^{\ddag}$ & -29.00 & 31.75 & \textbf{50.50}$^{\ddag}$ & +18.75 & 9.75 & \textbf{20.00}$^{\ddag}$ & +10.25 \\
& SWE-bench & 23.50 & \textbf{9.50}$^{\ddag}$ & -14.00 & 61.00 & \textbf{67.50}$^{\dag}$ & +6.50 & 15.50 & \textbf{23.00}$^{\dag}$ & +7.50 \\
& GAIA & 27.50 & \textbf{7.75}$^{\ddag}$ & -19.75 & 43.25 & \textbf{62.25}$^{\ddag}$ & +19.00 & 29.25 & \textbf{30.00} & +0.75 \\
\midrule
\multirow{3}{*}{\textbf{Mistral v0.3}} 
& Multi-Challenge & 42.25 & \textbf{40.50} & -1.75 & 56.00 & \textbf{57.50} & +1.50 & 1.75 & \textbf{2.00} & +0.25 \\
& SWE-bench & 51.75 & \textbf{43.25}$^{\dag}$ & -8.50 & 47.75 & \textbf{55.25}$^{\dag}$ & +7.50 & 0.50 & \textbf{1.50} & +1.00 \\
& GAIA & 46.75 & \textbf{37.00}$^{\ddag}$ & -9.75 & 51.00 & \textbf{62.00}$^{\ddag}$ & +11.00 & \textbf{2.25} & 1.00 & -1.25 \\
\bottomrule
\end{tabular}
\caption{Stance Breakdown across datasets. CED suppresses the \texttt{ADOPTED} stance. The data reveals distinct architectural behaviors upon recovering sovereignty: Gemma-2 transitions its reclaimed bandwidth into passive task focus (\texttt{IGNORED}), whereas Llama-3.1 leverages its independence to refute the adversarial swarm (\texttt{REJECTED}). Statistical significance of the CED shift vs. Standard Decoding is denoted by $^{*}$ ($p<0.05$), $^{\dag}$ ($p<0.01$), and $^{\ddag}$ ($p<0.001$).}
\label{tab:stances}
\vspace{-0.6cm}
\end{table*}

% \begin{figure}[t] % [t] is often necessary for two-column figures
% \vspace{-0.5cm}
%   \centering
%   \includegraphics[width=0.7\linewidth]{Figures/compliance_capability.png}
%   \caption{Decoupling Compliance and Capability.}
%   \label{fig:decouple}
% \vspace{-0.6cm}
% \end{figure}

\subsection{Parameterization of the CED Engine}
\label{subsec:results_optimization}

Our hyperparameter grid search identifies the boundary conditions required to safely suppress multi-agent sycophancy without degrading structural syntax.\\
\textbf{The Optimal Penalty ($\alpha$): }
As illustrated in Figure \ref{fig:alpha}, high penalties in log-space act as a destructive intervention, corrupting the reasoning path by distorting grammatical structure. Upon transitioning the mathematical penalty to the absolute probability space via the zero-bounded clamp, the magnitude required to successfully attenuate sycophantic tokens increases drastically. We found that $\alpha = 20$ represents the optimal plateau; lower values ($\alpha \le 5.0$) fail to overcome the dense log-probability compression of the conformity bias, while extreme values ($\alpha \ge 50.0$) begin to induce minor accuracy degradation. This selected $\alpha$ successfully satisfies our bivariate optimization constraint: reducing sycophancy ($\Delta \text{Loafing} \le 0$) while bounding standard accuracy degradation to a maximum of 5\% ($\Delta \text{Accuracy} \ge -0.05$). \\
\textbf{The Adaptive Plausibility Constraint ($k$): }
We evaluate continuous probability-mass thresholds against varying discrete thresholds for the top-$k$ truncation mask. Because models assign an overwhelming majority of the probability mass to the swarm's decoy, continuous confidence limits fail to restrict the vocabulary adequately. Therefore, we discard continuous margins in favor of a discrete boundary, identifying $k=3$ as the optimal truncation mask. 
This narrow boundary is sufficient because the model's standard probability distribution is highly peaked under adversarial load; it perfectly protects structural syntax (such as intermediate <thought> tags) without crippling generation diversity, while preserving sufficient bandwidth for the CED penalty to intervene.\\
\textbf{Evaluating Grammatical Safety Masks: }
Our ablation of mathematical masking strategies used in standard CD reveals critical flaws in standard confidence-based exemptions. We observe that the \textit{Dynamic Confidence Mask} backfired, as models exhibit unusually high confidence in sycophantic errors under social load; consequently, the mask inadvertently shields the exact adversarial tokens it is designed to penalize. Similarly, applying a \textit{Divergence Margin Mask} fails to reliably separate grammatical noise from sycophantic drift, often neutralizing the necessary penalty. The probability clamp, $\max(\Delta, 0)$, emerges as a mathematically sound solution to prevent artificial boosting of out-of-distribution artifacts.

\vspace{-0.3cm}
\subsection{Recovery of Agentic Sovereignty}
\label{subsec:recovery}
\vspace{-0.1cm}

Figure \ref{fig:accuracy_loafing} and Table \ref{tab:main_results} present the efficacy of CED across the evaluation corpus. The empirical data demonstrates that CED application mathematically suppresses the sycophantic loafing rate across structurally distinct architectures and task domains. For susceptible models, the suppression of social conformity drives statistically significant recoveries in objective task performance ($p < 0.001$).
Gemma-2 exhibits the most pronounced behavioral correction, demonstrating a near-doubling of accuracy across complex reasoning environments. On SWE-bench, Gemma-2 recovers from a baseline accuracy of 33.75\% under Standard Decoding to 60.25\% with CED, while its sycophantic loafing simultaneously drops by 28.00\% absolute (from 66.25\% to 38.25\%). This robust recovery scales proportionately to GAIA, where CED yields a 30.75\% absolute accuracy increase (from 35.50\% to 66.25\%) alongside a 33.00\% absolute reduction in cognitive loafing (from 64.25\% to 31.25\%).
Llama-3.1 demonstrates significant structural corrections across diverse domains, validating the domain invariance of the intervention. In low-entropy conversational environments (Multi-Challenge), CED halves Llama-3.1's loafing rate from 58.50\% to 29.50\%, driving a highly significant ($p < 0.001$) accuracy recovery from 38.75\% to 52.00\%. Conversely, on complex reasoning tasks (GAIA), it suppresses loafing from an initial 27.50\% down to a mere 7.75\%. Despite this 19.75\% drop in sycophancy, the accuracy gains remain mathematically marginal, shifting only from 55.00\% to 57.50\%. 
Finally, while Mistral v0.3 exhibits more modest accuracy improvements, the statistical suppression of its loafing behavior confirms CED's universal efficacy as an architectural filter. On SWE-bench, CED significantly reduces Mistral's loafing rate from 51.75\% to 43.25\% ($p < 0.01$), and on GAIA from 46.75\% to 37.00\% ($p < 0.001$). This systematically shows that CED successfully neutralizes peer pressure even when the model's underlying latent reasoning capacity prevents corresponding downstream accuracy gains.

\vspace{-0.2cm}
\subsection{Architectural Bias and the Lead Anchor Effect}\label{subsec:flattening}

A model's susceptibility to peer pressure depends heavily on the swarm's architectural makeup and speaking sequence. The Standard Decoding heatmap (Figure \ref{fig:heatmap}, left) provides two key visual evidences of this bias: \textbf{(1) Homogeneous Kinship:} Models exhibit peak sycophancy when pressured by peers of their own architecture, explicitly visible as the dark column in the baseline heatmap. This aligns with the concept of
"Architectural Tribalism" and also validates the consensus paradox, both established in \cite{shehata2026inversewisdom}
\textbf{(2) The Lead Anchor Effect:} Permuting the speaking order of \textit{Heterogeneous Strangers} dynamically alters the loafing rate across the heterogeneous columns, confirming conformity is positionally dependent, supporting prior findings on agentic bystander dynamics \cite{shehata2026bystander}.
CED flattens this variance. As shown in the post-intervention heatmap (Figure \ref{fig:heatmap}, right), applying the asymmetric probability clamp neutralizes both the tribal and positional biases. This acts as an equalizer, suppressing the loafing rate to a uniformly low baseline regardless of the swarm's composition or sequence.

% \subsection{Stance Transitions: Task Focus vs. Refutation}
% \label{subsec:stances}

% To show that CED does not achieve accuracy gains through stochastic guessing, Figure \ref{fig:stances} and Table \ref{tab:stances} map the behavioral shifts in the models' reasoning trajectories. The data confirms a universal suppression of the \texttt{ADOPTED} stance. Crucially, we observe distinct architectural behaviors in how models process this reclaimed independence. 
% Gemma-2 transitions its reclaimed bandwidth almost exclusively into passive task focus (\texttt{IGNORED}). For example, on the Multi-Challenge dataset, Gemma-2's \texttt{ADOPTED} trajectories drop from 264 to 170, which is perfectly absorbed by the \texttt{IGNORED} stance rising from 134 to 229, while its \texttt{REJECTED} stance remains negligible. Conversely, Llama-3.1 leverages its independence to actively refute the adversarial swarm (\texttt{REJECTED}). On SWE-bench, Llama-3.1 increases its \texttt{REJECTED} trajectories from 62 to 92 under CED, demonstrating an underlying capacity to explicitly identify and correct peer logical errors once freed from the sycophantic attractor.

\vspace{-0.1cm}
\subsection{Stance Transitions: Sycophancy Mitigation}

Figure \ref{fig:stances} and Table \ref{tab:stances} demonstrate that CED consistently suppresses the sycophantic \textsc{Adopted} stance across all benchmarks. Examining the percentage shifts reveals distinct architectural behaviors upon recovering sovereignty:
\textbf{(1) Generic Sycophancy Reduction:} Applying CED significantly reduces the \textsc{Adopted} stance. For instance, Llama-3.1's baseline compliance on the Multi-Challenge dataset drops by nearly half, from 58.50\% to 29.50\%, and plummets to just 7.75\% on the GAIA benchmark.
\textbf{(2) Passive Task Focus (Gemma-2):} When freed from peer pressure, Gemma-2 transitions its reclaimed bandwidth into passive task focus (the \textsc{Ignored} stance). This is most evident on the GAIA dataset, where a 33.00\% drop in its \textsc{Adopted} stance is mirrored exactly by a 33.00\% increase in its \textsc{Ignored} stance, with outright rejections remaining negligible (0.25\%).
\textbf{(3) Active Refutation (Llama-3.1):} Conversely, Llama-3.1 leverages its independence to actively refute the adversarial swarm. On the Multi-Challenge dataset, its \textsc{Rejected} stance sees a notable +10.25\% increase alongside a +18.75\% rise in the \textsc{Ignored} stance, demonstrating a much more confrontational recovery of reasoning sovereignty.

\vspace{-0.1cm}
\subsection{Decoupling Compliance and Capability}
\label{subsec:decoupling}

Figure \ref{fig:decoupling} decouples compliance reduction from objective performance to validate CED
% 's mathematical integrity
. While CED produces massive accuracy gains in Gemma-2 and Llama-3.1, its effect on Mistral v0.3 is primarily behavioral. On SWE-bench in Table \ref{tab:main_results}, CED significantly reduces Mistral's loafing rate from 51.75\% to 43.25\% ($p < 0.01$), but its objective accuracy sees only marginal gains, shifting from 25.25\% to 26.00\%.
We hypothesize this lack of strong accuracy recovery correlates with Mistral possessing the lowest parameter number among the tested architectures, inherently limiting its latent reasoning bandwidth.
This ablation shows that CED acts as a universal sycophancy suppressor across tested architectures. However, it cannot artificially inject latent reasoning capabilities if the base model lacks them. When a model lacks the capacity to solve the multi-hop semantic background, CED prevents it from sycophantically cheating, confirming that it restores epistemic sovereignty without inflating competence.

\section{Discussion, Limitations and Future Work}

CED neutralizes both architectural tribalism and positional bias, acting as a zero-shot sycophancy suppressor without fine-tuning. Our findings illuminate avenues for future work.
\textbf{(1) Computational Trade-offs:}
Like standard CD \cite{li-etal-2023-contrastive}, CED relies on a dual forward-pass, imposing a $2\times$ compute latency compared to standard generation. However, CED is superior in memory efficiency, eliminating the need to load auxiliary amateur models. While latency increases, this trade-off is highly justified in complex, high-stakes domains where zero-shot accuracy recovery outweighs latency, remaining vastly more resource-efficient than continuous preference fine-tuning. Though we mitigate pre-fill overhead by caching Key-Value states, future research should explore speculative decoding to bypass the dual-pass requirement for high-confidence tokens.
% CED design relies on a dual forward-pass to calculate the contrastive divergence. While this imposes a high inference cost, this trade-off is highly justified in complex, high-stakes domains where zero-shot accuracy recovery outweighs latency. While we mitigate the pre-fill overhead by caching past key and values during the initial prompt processing, the autoregressive generation phase still carries a computational cost. Future research should explore integrating speculative decoding frameworks to bypass the dual-pass requirement for high-confidence tokens.
% , triggering CED exclusively when the entropy of the base distribution spikes.
\textbf{(2) Architectural Accessibility:} Because CED requires direct manipulation of pre-softmax log-probabilities, its implementation is restricted to open-weight architectures. Translating contrastive epistemic interventions to closed-API systems—perhaps via guided prompt injection rather than logit manipulation—represents a next step for scaling sycophancy suppression.
\textbf{(3) Dynamic vs. Static Swarms:}
% To ensure strict isolation and measure pure conformity bias without the confounding variable of a live peer's persuasiveness, our evaluation utilizes static adversarial dialogue. Future work will deploy the CED engine within live multi-agent swarms to study its impact on long-term consensus convergence.
To ensure isolation and measure conformity bias without the confounding variable of a live peer's persuasiveness, our evaluation utilizes static adversarial dialogue. While live interactions ultimately resolve into a static context prefix for the terminal synthesizer agent, static evals prevent uncontrolled linguistic drift from confounding the measurement. Future work can deploy CED within live MAS.
% to study its impact on long-term consensus convergence.
\textbf{(4) Capability Bounds and Knowledge Injection:} As demonstrated by Mistral v0.3, CED recovers epistemic sovereignty but cannot inject latent reasoning capabilities if the base model lacks them.
This establishes CED as a behavioral filter rather than a reasoning enhancer. Future directions may explore coupling CED with retrieval-augmented generation (RAG) \cite{gema-etal-2025-decore, corallo-papotti-2026-parallel} to supply the required multi-hop knowledge when the base model lacks the latent capability to refute an adversarial consensus.

\vspace{-0.1cm}
\section{Related Works}
\label{sec:related_works}

\textbf{(1) Mitigating Conformity and Strategic Deception.} 
LLMs exhibit sycophancy and strategic deception due to RLHF \cite{sharma2024towards, 3600270.3602281, perez2022discovering}. Prior mitigations rely on resource-intensive alignment strategies, such as preference fine-tuning via synthetic data \cite{pang2024iterative, wei2025simple}, representation engineering via activation steering \cite{zou2023representation, rimsky-etal-2024-steering}, or structural loss regularization \cite{zhang2026stable}. While effective, these methods pose severe scalability bottlenecks (training overhead, degradation of general capabilities, etc.) for MAS. We diverge by proposing CED, which isolates and neutralizes conformity strictly through zero-shot, inference-time logit manipulation.
\textbf{(2) Contrastive Decoding:} 
Our work builds upon CD \cite{li-etal-2023-contrastive}, a logit-subtraction technique adapted by concurrent works to mitigate hallucinations via network layers (DoLa) \cite{chuang2024dola} and enforce retrieval grounding (Context-Aware Decoding) \cite{shi-etal-2024-trusting}. 
While foundational CD relies on continuous linear subtraction and recent variants \cite{chen2026beyond} explore attention-level contrast, they fail under the extreme multipliers to override MAS consensus. To our knowledge, CED is the first CD framework targeting multi-agent sycophancy
% . 
% We replace linear subtraction with an asymmetric, zero-bounded probability clamp and a discrete top-k mask, 
preventing the artificial token boosting and grammatical collapse seen in prior works under heavy penalties.

\vspace{-0.1cm}
\section{Conclusion}

To mitigate sycophancy, we introduce CED, a zero-shot intervention that replaces linear logit subtraction with an asymmetric probability clamp and discrete top-$k$ mask. Evaluated across 7,200 trajectories, CED acts as a positionally invariant equalizer, suppressing cognitive loafing by up to 33.00\% and yielding accuracy recoveries up to 30.75\%. CED establishes a training-free behavioral filter that decouples compliance from capability, restoring agentic sovereignty without inflating the competence of capacity-limited architectures.

\bibliography{aaai2027}


\end{document}